\documentclass{article}
\usepackage{arxiv}

\publisedin{\scriptsize Preprint of article published in \textit{AI 2019: Advances in Artificial Intelligence}. 
    See \url{https://doi.org/10.1007/978-3-030-35288-2_22}}

\usepackage{graphicx}
\usepackage{amsmath}

\usepackage{booktabs}

\usepackage{nicefrac}

\usepackage{tikz}
\usetikzlibrary{decorations.pathreplacing, positioning, shapes, arrows.meta}
\tikzset{>=latex}

\usepackage{subcaption}

\usepackage{layouts}

\DeclareSymbolFont{rsfso}{U}{rsfso}{m}{n}
\DeclareSymbolFontAlphabet{\mathscr}{rsfso}

\usepackage{hyperref}
\hypersetup
{
	pdftitle = {Sequence-to-Sequence Imputation of Missing Sensor Data},
	pdfauthor = {Joel Janek Dabrowski},
}

\begin{document}
	\title{Sequence-to-Sequence Imputation of Missing Sensor Data}
	%
	%
	\author{
        Joel Janek Dabrowski \\ 
        Data61, CSIRO, Australia, \\ 
        \texttt{Joel.Dabrowski@data61.csiro.au} 
        \And
		Ashfaqur Rahman \\ 
        Data61, CSIRO, Australia, \\
        \texttt{Ashfaqur.Rahman@data61.csiro.au}
    }
	\maketitle              
	\begin{abstract}
		Although the sequence-to-sequence (encoder-decoder) model is considered the state-of-the-art in deep learning sequence models,
		there is little research into using this model for recovering missing sensor data. The key challenge is that the missing sensor data problem typically comprises \textit{three} sequences (a sequence of observed samples, followed by a sequence of missing samples, followed by another sequence of observed samples) whereas, the sequence-to-sequence model only considers \textit{two} sequences (an input sequence and an output sequence). We address this problem by formulating a sequence-to-sequence in a novel way. A forward RNN encodes the data observed before the missing sequence and a backward RNN encodes the data observed after the missing sequence. A decoder decodes the two encoders in a novel way to predict the missing data. We demonstrate that this model produces the lowest errors in 12\% more cases than the current state-of-the-art.
		
	\end{abstract}
	

\section{Introduction and Related Work}

From smart cities \cite{Arasteh2016iot} to personalised body sensor networks \cite{Poon2015Body}, sensor data is becoming ubiquitous. This has been fuelled by the rise of the internet of things (IOT), smart sensor networks, and low-cost sensors. Such technologies are however imperfect and their failure may result in \textit{missing data}. Sensors may fail due to hardware or software failure. Communication networks can break down due to low signal level, network congestion, packet collision, limited memory capacity, or communication node failures \cite{yick2008wireless}. Even if sensors and communications prevail, missing data may result from scheduled outages such as maintenance and upgrade routines.

When a data-driven model (such as a machine learning model) uses sensor data for prediction, missing data introduces various challenges in parameterising or training the model. This is especially problematic when the temporal structure of the data is important to the model. To address this problem, various methods for imputing or interpolating the missing data have been proposed in the literature.


The Recurrent Neural Network (RNN) has been shown to perform well for missing data recovery \cite{che2018recurrent,yoon2018deep,cao2018brits,Shen2018End}. However, there is little research into using the sequence-to-sequence (encoder-decoder) model \cite{Sutskever2014Sequence}, despite it being considered as a state-of-the-art model in deep learning sequence modelling. The key challenge in applying this model to missing data is that it is designed to use a \textit{single} input sequence to predict some output sequence. However, the missing data problem can be considered to have \textit{two} input sequences that are separated by the missing data. That is, relative to the missing data, the model must take into account data that is observed before and after the missing data sequence.

We propose a novel sequence-to-sequence model that incorporates the data before and after a missing data sequence to predict the missing values of that sequence. For this, two encoders are used: one propagating in the positive time-direction, and one propagating in the negative-time direction. These two encoders feed into a decoder that naturally combines the encoded forward and backward encoders to provide an accurate prediction of the missing data sequence. A key feature of the sequence-to-sequence model is that it can handle arbitrary length input and output sequences. Our key contributions are:
\begin{enumerate}
	\item The proposed decoder architecture is novel in the way that it merges information from two encoders.
	\item We introduce a novel approach to scaling a forward and backward RNN within the decoder according to their proximity to observed data.
	\item We demonstrate results which show that our model outperforms the current state-of-the-art methods on several datasets.
\end{enumerate}

The proposed model is particularly applicable in problems where there is no neighbouring data available for imputing across variables at each sequence step. The recovery of the missing data must be determined from temporal information alone. These include univariate problems or multivariate problems where sequences of data are missing across \textit{all} measured variables at the same time. This typically occurs when there is a central system failure, such as the failure of a multi-parameter sensor, the failure of a central communications node in a star-network, or a scheduled outage across a system. 


\section{Related Work}


Various models such as MICE \cite{Buuren2000Multivariate} and ST-MVL \cite{yi2016st} have been proposed for missing data recovery. We however focus on RNNs, as these are considered to be the state-of-the-art in many missing data recovery applications. Various forms of the RNN have been tested for data imputation. Che et. al. \cite{che2018recurrent} use the Gated Recurrent Unit with trainable decays (GRU-D) model for recovering missing data. The decay rates exponentially reduce importance of predictions that are distant from observations. The model however does not consider samples that occur after the missing data sequence.

The M-RNN \cite{yoon2018deep} uses a bidirectional neural network for imputation. The model is a multi-directional RNN that considers temporal information as well as information across sensors to recover missing data. This model thus relies on a subset of data to be available at any time. 

The RITS and BRITS \cite{cao2018brits} model use a RNN to perform one-step ahead forecasting and modelling over sequences. Compared with M-RNN, it trains output nodes with missing data as part of the network. Both this and a bidirectional RNN provide a means learn from data that lies before and after the missing data sequence. Additionally, they use trainable decays similar to the GRU-D. However, like the M-RNN, the RITS and BRITS models perform imputation by considering temporal information as well as information across sensors. Cao et. el. \cite{cao2018brits} do however propose the RITS-I and BRITS-I models as reduced versions of RITS and BRITS which exclude the mechanism used to perform predictions across sensors. These reduced models focus on temporal predictions and are thus used for comparison in this study.


The Iterative Imputing Model (IIM) \cite{zhou2018recover} uses a forward and backward RNN to encode information before and after the missing data. These RNNs could be considered to perform the task of the encoder in the sequence-to-sequence model. However, to predict the missing data, a predict-update loop (similar to the EM algorithm) is used iteratively impute each missing sample. This iterative process is computationally expensive and does not correspond with a decoder in the sequence-to-sequence model.

The SSIM model \cite{zhang2019ssim} is the first model to use the sequence-to-sequence approach for recovering missing data. To address the problem of including observations before and after the missing data, SSIM uses a forward and backward RNN together with a variable-length sliding window. A drawback of the model is that it has to ``learn'' that there is a difference between the observations before and after the missing data \cite{zhang2019ssim}.

Compared with GRU-D, BRITS, and M-RNN, our model uses the sequence-to-sequence approach, which is the state-of-the-art in applications such as natural language processing. Furthermore, we consider the problem where there is a complete set of data across all sensors or variables. The result is that data recovery is performed on temporal information alone. Compared with IIM, our model uses an arbitrary length decoder that does not require an iterative updating approach. Compared with SSIM, our model naturally stitches the observations before and after the missing data and is thus not required to learn that there is a difference between them. Furthermore, it does not require a variable sliding window to operate.


\section{Model}

\subsection{Architecture}

\begin{figure}[!t]
	\centering

\def\horisep{0.89cm}
\def\vertsep{0.65cm}

\begin{footnotesize}
\begin{tikzpicture}[->, >={Latex[length=1.5mm, width=1mm, black!40]}, draw=black!40, thick]
	\tikzstyle{lstmfw}=[fill=black!50,draw=black!50,minimum size=9pt,inner sep=0pt]
	\tikzstyle{lstmbw}=[fill=black!65,draw=black!65,minimum size=9pt,inner sep=0pt]
	\tikzstyle{output}=[circle, fill=black!35,draw=black!55,minimum size=9pt,inner sep=0pt]
	\tikzstyle{times}=[circle, draw=black!55, minimum size=9pt,inner sep=0pt]
	\tikzstyle{label}=[inner sep=1pt];
	\pgfsetshortenstart{1pt}
	\pgfsetshortenend{1pt}
	%
	
	\draw[rounded corners, draw=black!20] (-0.5*\horisep, -0.3cm) 
	rectangle (2.5*\horisep, 6*\vertsep+0.3cm) {};
	
	\draw[rounded corners, draw=black!20] (3.0*\horisep, -0.3cm) 
	rectangle (9.5*\horisep, 6*\vertsep+0.3cm) {};
	
	\draw[rounded corners, draw=black!20] (10*\horisep, -0.3cm) 
	rectangle (13*\horisep, 6*\vertsep+0.3cm) {};

	\node[label] (enc) at (1*\horisep,-0.6cm) {Forward Encoder};
	\node[label] (enc) at (6.25*\horisep,-0.6cm) {Decoder};
	\node[label] (enc) at (11.5*\horisep,-0.6cm) {Backward Encoder};
	
	\foreach \x in {0,...,2}
	{
		\node[label] (Lfwl\x) at (\x*\horisep, 0*\vertsep) {$x_{\x}$};
		\node[lstmfw] (Lfw\x) at (\x*\horisep, 1*\vertsep) {};
		\path (Lfwl\x) edge (Lfw\x);
		\ifnum \x>0
			\pgfmathtruncatemacro{\y}{\x-1}
			\path (Lfw\y) edge (Lfw\x);
		\fi
	}

	\foreach \x in {4,6,8}
	{
		\node[lstmfw] (Lfw\x) at (\x*\horisep, 1*\vertsep) {};
		\pgfmathtruncatemacro{\y}{\x-2}
		\path (Lfw\y) edge (Lfw\x);
		\ifnum \x>4
			\draw [dashed] (Lfw\y.north) to [out=60,in=240, looseness=1.2] (Lfw\x.south);
		\fi
	}
	\draw (Lfwl2.east) to [out=0,in=270, looseness=1.2] (Lfw4.south);

	\foreach \x in {12,11,10}
	{
		\pgfmathtruncatemacro{\idx}{\x-4}
		\node[label] (Lbwl\x) at (\x*\horisep + \horisep/2, 0*\vertsep) {$x_{\idx}$};
		\node[lstmbw] (Lbw\x) at (\x*\horisep + \horisep/2, 2*\vertsep) {};
		\path (Lbwl\x) edge (Lbw\x);
		\ifnum \x<8
			\pgfmathtruncatemacro{\y}{\x+1}
			\path (Lbw\y) edge (Lbw\x);
		\fi
	}

	\foreach \x in {8,6,4}
	{
		\node[lstmbw] (Lbw\x) at (\x*\horisep + \horisep, 2*\vertsep) {};
		\pgfmathtruncatemacro{\y}{\x+2}
		\path (Lbw\y) edge (Lbw\x);
		\ifnum \x<8			
			\draw [dashed] (Lbw\y.north) to [out=120,in=300, looseness=1.2] (Lbw\x.south);
		\fi
	}
	\draw (Lbwl10.west) to [out=180,in=270, looseness=1.2] (Lbw8.south);

	\foreach \x in {4,6,8}
	{
		\node[times] (Lfwt\x) at (\x*\horisep, 4*\vertsep) {$\times$};
		\node[times] (Lbwt\x) at (\x*\horisep + \horisep, 4*\vertsep) {$\times$};
		
		\path (Lfw\x.north) edge (Lfwt\x);
		\path (Lbw\x.north) edge (Lbwt\x);
		
		\pgfmathtruncatemacro{\idx}{\x/2+1}
		\node[label] (Lgfw\x) at (\x*\horisep - \horisep/2, 3*\vertsep) {$\gamma_\idx$};
		\node[label] (Lgbw\x) at (\x*\horisep + \horisep/2, 3*\vertsep) {$\gamma_\idx'$};
		
		\path (Lgfw\x.north) edge (Lfwt\x);
		\path (Lgbw\x.north) edge (Lbwt\x);
	}

	\foreach \x in {4,6,8}
	{
		\pgfmathtruncatemacro{\idx}{\x/2+1}
		\node[output] (L\x) at (\x*\horisep, 5*\vertsep) {};
		\node[label] (Lout\x) at (\x*\horisep, 6*\vertsep) {$\hat{x}_{\idx}$};
		\path (L\x) edge (Lout\x);
		\path (Lfwt\x.north) edge (L\x);
		\path (Lbwt\x.north) edge (L\x);
	}
	
\end{tikzpicture}
\end{footnotesize}
	\caption{Illustration of proposed sequence-to-sequence model for missing sensor data imputation. Square nodes denote LSTM cells and circular nodes denote linear output neurons. The circular nodes with the $\times$ operator denote element-wise multiplication with the scaling factors $\gamma_t$ and $\gamma_t'$. Observations are provided for $x_0, x_1, x_2$ and $x_6, x_7, x_8$. The values for $x_3, x_4, x_5$ are missing. The forward encoder encodes $x_0, x_1, x_2$ and the backward encoder encodes $x_6, x_7, x_8$. The decoder is a bidirectional LSTM that predicts $\hat{x}_3, \hat{x}_4, \hat{x}_5$. Each forward and backward LSTM cell in the decoder predicts the missing data and this prediction is input to the next RNN cell in the sequence as illustrated by the dashed arrows. The LSTMs in the decoder thus perform one-step-ahead forecasting.}
	\label{fig:fig_seq2seqImputation}
\end{figure}
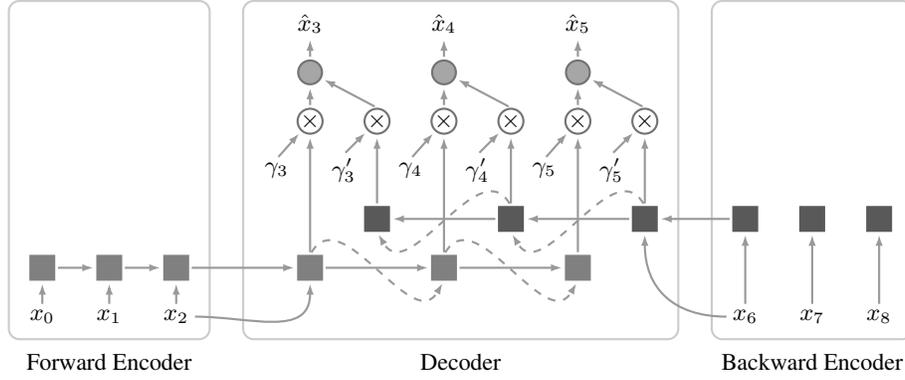

A sequence-to-sequence (encoder-decoder) model is proposed to recover missing time series data. As illustrated \figurename{~\ref{fig:fig_seq2seqImputation}}, the network comprises a forward encoder, a backward encoder, and a form of  bidirectional decoder. The network can be viewed as containing two traditional sequence-to-sequence models \cite{Sutskever2014Sequence}, one in the forward direction, and one in the backward direction. The outputs of the forward and backward RNN cells in the decoder are scaled and merged together in a final output layer in the form of a Multilayered Perceptron (MLP).

The forward and backward decoder RNNs operate by performing one-step-ahead predictions. The prediction of the previous RNN cell is fed to the input of the current cell as illustrated by the dashed arrows in \figurename{~\ref{fig:fig_seq2seqImputation}}. In a regression problem, the prediction is performed using a MLP with a linear output layer and inputs given by the outputs of the corresponding RNN cell. The forward encoder predictions are denoted by $\hat{x}^\text{FW}_t$ and the backward encoder predictions are denoted by $\hat{x}^\text{BW}_t$. 

The additional outputs at the RNN level are required as all the final output layer's outputs are not available at each sequence step. For example, as illustrated in \figurename{~\ref{fig:fig_seq2seqImputation}}, computing $\hat{x}_4$ requires the output of the second forward RNN cell and the second backward RNN cell. If the final output layer outputs were fed to the next cell, $\hat{x}_3$ would be fed to the input of the forward RNN at index 4. However, $\hat{x}_3$ also requires the output of the third backward RNN cell, which is not available as the backward RNN has only been processed up to its second cell. To address this dilemma, the forward RNNs and the backward RNNs are first processed over the entire sequence with their local outputs. The results are then passed to the final output layer.

\subsection{Scaling Factors}
\begin{figure}[!t]
    \centering
    \includegraphics{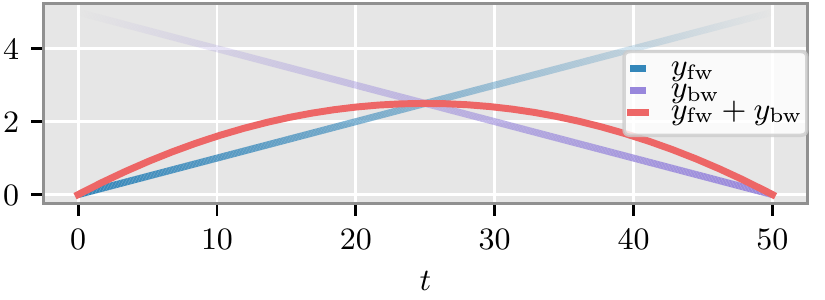}
    \caption{An illustration demonstrating the principle of the scaling factor approach using two arbitrary linear functions. The variable $y_\text{fw}$ decays with increasing $t$ (illustrated with a vanishing curve), whereas $y_\text{fw}$ decays with decreasing $t$. The prediction is the weighted combination of $y_\text{fw}$ and $y_\text{bw}$. In the proposed model, $y_{\text{bw}}$ is the output of the forward decoder RNN, $y_{\text{bw}}$ is the output of the backward decoder RNN, and the summation operation in $y_{\text{fw}} + y_{\text{bw}}$ is a nonlinear operation performed by the output layer of the model.}
    \label{fig:scalingIllustration}
\end{figure}
Before the outputs of the forward and backward decoder RNNs are merged together in the final output MLP, the RNN outputs are scaled with a scaling factor $\gamma_t$. In our novel approach, the scaling factor decays as predictions progress further from observed data. The forward RNN outputs in the decoder are scaled according to the linear function
\begin{align}
\gamma_t = 1 - \frac{t}{T}
\end{align}
where $T$ is the length of the missing data sequence and $t = \{1,\dots, T\}$ is the index of the missing data sequence samples. The backward RNN outputs in the decoder are scaled according to
\begin{align}
\gamma_t' = 1 - \gamma_t
\end{align}
Thus, at time $t=1$, the forward RNN output is scaled by a factor of $\gamma_1=1$. This factor decays to zero as $t$ increases. The opposite is true for the backward RNN, where it is scaled by a factor of $\gamma_T'=1$ at time $t=T$. This factor decays to zero as $t$ decreases. The result is that the forward decoder RNN is emphasised near the observations associated with the forward encoder and the backward decoder RNN is emphasised near the observations associated with the backward encoder. The principle of this process is illustrated in an example using linear functions in \figurename{~\ref{fig:scalingIllustration}}.

Scaling factors have been previously used in RNNs in \cite{che2018recurrent} and \cite{yoon2018deep}. These factors however decay exponentially and are integrated into the RNN network where they can be learned. In our approach, the scaling factors can be viewed as form of a ``forced'' attention mechanism that favours the RNN outputs that are nearest the observed data. Furthermore, the linear nature of the proposed scaling factors ensures a balanced weighting between the RNNs across the sequence such that $\gamma_t + \gamma_t'=1 ~ \forall t$.

\subsection{Backpropagation With Scaling Factors}

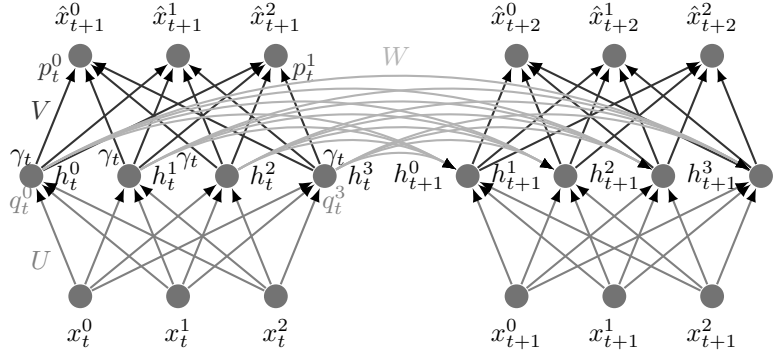
\begin{figure}[!t]
    \centering

\def\neuronsep{1.3cm}
\def\layersep{1.6cm}
\def\netsep{5.8cm}
\definecolor{mycolor1}{rgb}{0.5 ,  0.5,  0.5}
\definecolor{mycolor2}{rgb}{0.2,  0.2,  0.2}
\definecolor{mycolor3}{rgb}{0.7,  0.7 ,  0.7}
	
\begin{tikzpicture}[->,draw=black!50, thick, pin distance=0.3*\neuronsep]
	\tikzstyle{neuron}=[circle,fill=black!55,minimum size=9pt,inner sep=0pt]
	\tikzstyle{labl}=[];

	\foreach \name / \x in {0,...,2}
	\node[neuron,label=below:$x_t^{\x}$] (L4-\name) at (\x*\neuronsep+0.5*\neuronsep+1*\netsep,0*\layersep) {};
	\foreach \name / \x in {0,...,3}
	\node[neuron,label=right:$h_t^{\x}$] (L5-\name) at (\x*\neuronsep+1*\netsep,1*\layersep) {};
	\foreach \name / \x in {0,...,2}
	\node[neuron,label=above:$\hat{x}_{t+1}^{\x}$] (L6-\name) at (\x*\neuronsep+0.5*\neuronsep+1*\netsep,2*\layersep) {};
	\foreach \source in {0,...,2}
	\foreach \dest in {0,...,3}
	\path[draw=mycolor1] (L4-\source) edge (L5-\dest);
	\foreach \source in {0,...,3}
	\foreach \dest in {0,...,2}
	\path[draw=mycolor2] (L5-\source) edge (L6-\dest);
	
    \node[mycolor1] () at (-0.1*\neuronsep+\netsep,0.8*\layersep) {$q_t^0$};
    \node[mycolor1] () at (3.1*\neuronsep+\netsep,0.8*\layersep) {$q_t^3$};
    \node[mycolor2] () at (0.2*\neuronsep+\netsep,1.9*\layersep) {$p_t^0$};
    \node[mycolor2] () at (2.8*\neuronsep+\netsep,1.9*\layersep) {$p_t^1$};
    
	\node[mycolor1] () at (0.1*\neuronsep+\netsep,0.3*\layersep) {$U$};
	\node[mycolor2] () at (0.1*\neuronsep+\netsep,1.55*\layersep) {$V$};
	\node[mycolor3] () at (1.5*\netsep+1.5*\neuronsep,2.0*\layersep) {$W$};
    
    \node[] () at (-0.1*\neuronsep+\netsep,1.15*\layersep) {$\gamma_t$};
    \node[] () at (0.80*\neuronsep+\netsep,1.15*\layersep) {$\gamma_t$};
    \node[] () at (1.6*\neuronsep+\netsep,1.13*\layersep) {$\gamma_t$};
    \node[] () at (3.1*\neuronsep+\netsep,1.15*\layersep) {$\gamma_t$};
	
	
	\foreach \name / \x in {0,...,2}
	\node[neuron,label=below:$x_{t+1}^{\x}$] (L7-\name) at (\x*\neuronsep+0.5*\neuronsep+2*\netsep,0*\layersep) {};
	\foreach \name / \x in {0,...,3}
	\node[neuron,label=left:$h_{t+1}^{\x}$] (L8-\name) at (\x*\neuronsep+2*\netsep,1*\layersep) {};
	\foreach \name / \x in {0,...,2}
	\node[neuron,label=above:$\hat{x}_{t+2}^{\x}$] (L9-\name) at (\x*\neuronsep+0.5*\neuronsep+2*\netsep,2*\layersep) {};
	\foreach \source in {0,...,2}
	\foreach \dest in {0,...,3}
	\path[draw=mycolor1] (L7-\source) edge (L8-\dest);
	\foreach \source in {0,...,3}
	\foreach \dest in {0,...,2}
	\path[draw=mycolor2] (L8-\source) edge (L9-\dest);
	
	\foreach \source in {0,...,3}
	\foreach \dest in {0,...,3}
	\path[draw=mycolor3] (L5-\source) edge[out=30, in=150, looseness=0.90] (L8-\dest);
	
	
\end{tikzpicture}
    \caption{An illustration of a forward RNN and the output layer in the decoder for the discussion on backpropagation with the scaling factor.}
    \label{fig:fig_rnn2}
\end{figure}

As the scaling factor scales the predictions, it also scales the derivatives used in backpropagation. This is due to $\gamma_t$ being a fixed constant. 
For example, consider the forward decoder RNN (in the form of an Elman network for illustrative purposes) with the output layer as illustrated in \figurename{~\ref{fig:fig_rnn2}}. The scaling factor $\gamma_t$ is applied to each output $h_t^j$. The variable $p^{k}_{t}$ is linear combination of inputs at output neuron $k$, and $q^j_{t}$ is the linear combination of inputs at hidden neuron $j$. The weight matrices $U$, $W$, and $V$ are associated with the input-to-hidden, hidden-to-hidden, and hidden-to-output connections respectively.

Following the backpropagation derivation, the derivative of the cost with respect to the weight $u_{ij}$ connecting the $i^\text{th}$ input to the $j^\text{th}$ RNN hidden node is given by
\begin{align*}
\frac{\partial \mathscr{L}}{\partial u_{ij}} 
=\Bigg( 
\underbrace{ 
    \sum_{k=1}^{n_{o}}
    \frac{\partial \mathscr{L}}{\partial p^{k}_{t}} 
    \frac{\partial p^{k}_{t}}{\partial h^{j}_{t}}
}_{\text{hidden to output}}
+ 
\underbrace{ 
    \sum_{k=1}^{n_{h}}
    \frac{\partial \mathscr{L}}{\partial q^{k}_{t+1}} 
    \frac{\partial q^{k}_{t+1}}{\partial h^{j}_{t}}
}_{\text{hidden to hidden}}
\Bigg)
\underbrace{ 
\frac{\partial h^{j}_{t}}{\partial q^j_{t}}
\frac{\partial q^j_{t}}{\partial u^{ij}}.
}_{\text{input to hidden}}
\end{align*}
where $n_o$ is the number of output units and $n_h$ is the number of hidden units. The scaling factor affects the link between the hidden layer outputs $h_t^j$ and the output layer linear combination $p_t^k$. This corresponds to the second factor in the first term. The derivative of this term is computed as
\begin{align*}
\frac{\partial p^{k}_{t}}{\partial h^{j}_{t}} 
&= \frac{\partial}{\partial h^{j}_{t}} \sum_j  v_{jk} (\gamma_t h_t^j) + b_{vk} \\
&= \gamma_t v_{jk}
\end{align*}
where $b_{vk}$ is a bias. The scaling factor thus affects the derivatives passed back from the outputs to the hidden layers. The result is that, similar to the scaling of the predictions, the backpropagated errors are scaled to emphasise the RNN cells that are near the corresponding encoders. The scaling is thus incorporated into the learning process.

\subsection{Output Layer}

The scaled forward and backward decoder RNN outputs are passed to a MLP which predicts the missing data. The prediction provided by this output layer at time $t$ is denoted by $\hat{x}_t$. With linear outputs producing the predictions $\hat{x}_t$, $\hat{x}_t^\text{FW}$, and $\hat{x}_t^\text{BW}$, the cost function is given by
\begin{align}
\label{eq:loss}
\text{loss}
= \frac{1}{T} \sum_t \left( \mathscr{L}(x_t, \hat{x}_t)
 + \mathscr{L} (x_t, \hat{x}_t^\text{FW})
 + \mathscr{L} (x_t, \hat{x}_t^\text{BW}) \right)
\end{align}
where $x_t$ is the ground truth value for the missing sample at time $t$ and $\mathscr{L}()$ is the mean squared error loss function (for the regression case).


\section{Experiments}

Several freely-available datasets are used to evaluate and compare the proposed model. The PM2.5 air quality dataset (from 2014-2015) is used as it is become a benchmark used in several previous studies such as \cite{yi2016st}, \cite{cao2018brits}, and \cite{zhou2018recover}. Note that, imputations are made across time \textit{and} across sensors in these studies. However, in the current study, imputations are made across time only. In addition to this dataset, the Metro Interstate Traffic Volume dataset, the Birmingham Parking dataset \cite{stolfi2017predicting}, 
and the Beijing PM2.5 Air Quality dataset (from 2010-2014) \cite{Xuan2015Assessing} are used. These datasets are freely available from the UCI Machine Learning Repository\footnote{\url{https://archive.ics.uci.edu/ml/index.php}}. 

For the PM2.5 dataset, the PM2.5 data for sensor 001001 is used. In Traffic dataset, Temperature and traffic volume are used. Each parking area provides a unique variable in the Parking dataset. Finally, the Dew point, temperature, and pressure variables are used in the AirQuality dataset.


The Mean Absolute Error (MAE) and the Mean Relative Error (MRE) are used as performance metrics. These are given by \cite{zhou2018recover,yi2016st,cao2018brits}.
\begin{align*}
& \text{MAE} = \frac{1}{N} \sum_i^N | x_{t_i} - \hat{x}_{t_i} |,
& \text{MRE} = \frac{\sum_i^N |x_{t_i} - \hat{x}_{t_i}|}{\sum_i^N |x_{t_i}|}
\end{align*}
where $N$ is the total number of observations.

The proposed  model results are compared with results from the RITS-I \cite{cao2018brits}, BRITS-I \cite{cao2018brits}, and the sequence-to-sequence \cite{Sutskever2014Sequence} models. In all models, 64 hidden units are used in the Long-Short Term Memory (LSTM) \cite{hochreiter1997long} RNN. A linear layer is used in the final output layer. All models are trained using the standard backpropagation approach to minimise (\ref{eq:loss}) with the Adam optimisation algorithm \cite{kingma2014adam}. Early stopping is used to avoid overfitting in the datasets.

The dataset is split into a test and training set such that the last 80\% of the dataset is used as a test set. Training and test samples are extracted using a sliding window that is slid across the datasets. Each extracted window is split into a sequence of missing values, a sequence of observed values preceding the missing values, and a sequence of observed values following the missing values. The models are implemented in PyTorch and trained on Dual Xeon 14-core E5-2690 v4 Compute Nodes.


\section{Results and Discussion}

\begin{figure}[!t]
    \centering
    \begin{subfigure}[b]{2.4in}
        \includegraphics[width=\textwidth]{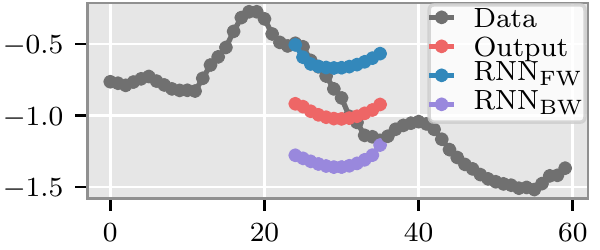}
        \caption{Without scaling factor}
        \label{fig:scalingWithoutGamma}
    \end{subfigure}
    \begin{subfigure}[b]{2.4in}
        \includegraphics[width=\textwidth]{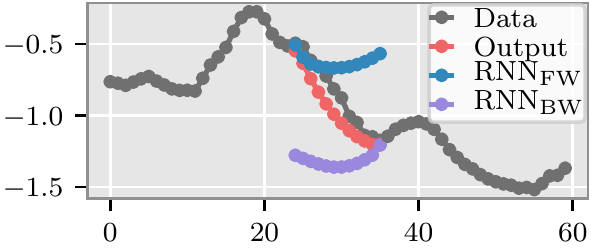}
        \caption{With scaling factor}
        \label{fig:scalingWithGamma}
    \end{subfigure}
    \caption{Demonstration of the scaling factor operation on a Traffic dataset sample. The scaling factor emphasises the forward decoder RNN at the beginning of the prediction and it emphasises the backward decoder RNN at the end of the prediction. The result is a more accurate prediction.}
    \label{fig:scalingFactor}
\end{figure}

To demonstrate the scaling factor, a prediction from a Traffic dataset sample is presented. The predictions of the forward decoder RNN, the backward decoder RNN, and the model output are plotted in \figurename{~\ref{fig:scalingFactor}}. The forward and backward RNNs produce significantly differing predictions. If the scaling factor is excluded from the model, the prediction is similar to the average of the forward and backward sequences. As both of these predictions deviate from the ground truth, this final prediction is inaccurate. By including the scaling into the model, the prediction is shifted towards the observed data points, providing a more accurate result.

\begin{figure}[!t]
    \centering
    \includegraphics{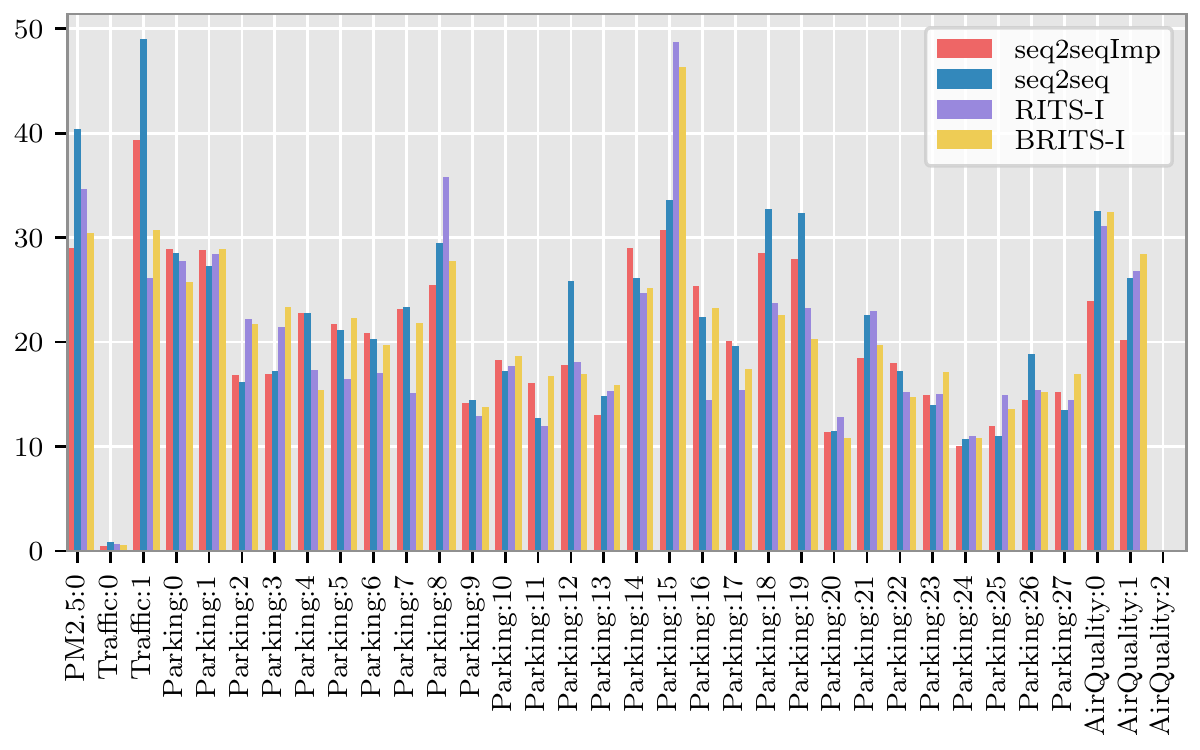}
    \caption{MRE\% results. See Table \ref{table:results_mae} for detailed MAE results. (The AirQuality results are not visible due to their scale. Refer to Table \ref{table:results_mae}).}
    \label{fig:maeBarChart}
\end{figure}

Table \ref{table:results_mae} lists the MAEs and \figurename{~\ref{fig:maeBarChart}} plots the MREs for the set of models and datasets. The Traffic dataset label indexes index the temperature and traffic volume variables in the dataset. The Parking dataset label indexes index the various parking areas. Finally, the AirQuality dataset label indexes index the dew point, temperature, and pressure variables. 
For reference, the dataset ranges are included in Table \ref{table:results_mae}. In figures and tables, the proposed model is denoted by seq2seqImp and the sequence-to-sequence model is denoted by seq2seq.

\begin{figure}[!t]
    \centering
    \includegraphics{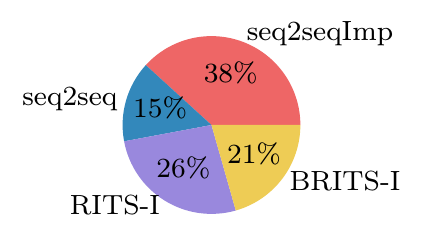}
    \caption{Pie chart indicating the share over which models produce optimal results.}
    \label{fig:pieChart}
\end{figure}

The share of optimal MAEs is presented as a pie chart in \figurename{~\ref{fig:pieChart}}. Overall, the proposed model has the highest share with 38\% of the lowest MAE results and is 12\% higher than the other models. The sequence-to-sequence has the smallest share with 15\% lowest errors. This is expected as the model is only provided with data prior to the missing data sequence. The other models are provided with data before and after the missing data sequence.

In the PM2.5 and AirQuality datasets, the proposed model produces significantly lower errors than the other models. For example, considering the proposed model produces MAEs that are a third lower than the competing models. The RITS-I model has the majority of its lowest errors in the Parking dataset. The model is thus well suited to this dataset. 

To provide an aggregated representation of the results, Borda counts are used to rank the models through voting. A Borda count ranks a set of $N$ models with integers $(1, \dots, N)$ such that the model with the highest error is assigned a value of 1 and the model with the lowest error is assigned a value of $N$. The sum of Borda counts for the models over all datasets are presented in Table \ref{table:results_borda}. The results show that the proposed model is voted as the highest ranked model.

\begin{footnotesize}
    \begin{table}[]
        \centering
        \caption{MAE on the datasets for the various models. The proposed model is dented by seq2seqImp. The forward decoder RNN prediction errors and the backward decoder RNN in the proposed model are included as $\text{RNN}_\text{FW}$ and $\text{RNN}_\text{BW}$ respectively. The sequence-to-sequence model is denoted by seq2seq.}
        \setlength{\tabcolsep}{1pt}
        \begin{tabular}{lccccccccc}
            \toprule
            & Range & seq2seqImp & $\text{RNN}_\text{FW}$ & $\text{RNN}_\text{BW}$ & seq2seq & RITS-I & BRITS-I \\
            \midrule
            PM2.5:0 & [3,429] & \textbf{11.13} & 16.27 & 15.50 & 16.95 & 15.50 & 13.92 \\
            Traffic:0 & [0,308] & \textbf{1.45} & 2.32 & 2.28 & 2.35 & 1.85 & 1.51 \\
            Traffic:1 & [0,7280] & 832.33 & 1027.22 & 1111.64 & 1021.09 & \textbf{621.72} & 682.48 \\
            Parking:0 & [20,492] & 55.82 & 65.82 & 66.52 & 51.25 & 54.54 & \textbf{49.70} \\
            Parking:1 & [0,320] & 36.45 & 39.06 & 42.87 & \textbf{31.54} & 33.71 & 33.95 \\
            Parking:2 & [68,821] & 106.73 & 143.12 & 127.77 & \textbf{106.05} & 143.84 & 139.90 \\
            Parking:3 & [39,402] & \textbf{56.68} & 65.11 & 61.18 & 58.41 & 72.29 & 78.57 \\
            Parking:4 & [0,1013] & 146.85 & 163.61 & 188.61 & 146.69 & 110.33 & \textbf{99.68} \\
            Parking:5 & [25,1197] & 136.90 & 158.31 & 211.23 & 133.06 & \textbf{105.68} & 142.88 \\
            Parking:6 & [15,612] & 53.61 & 59.70 & 78.22 & 50.89 & \textbf{43.56} & 50.70 \\
            Parking:7 & [30,470] & 50.17 & 61.18 & 80.78 & 54.91 & \textbf{38.96} & 41.62 \\
            Parking:8 & [2,220] & \textbf{38.77} & 51.53 & 42.93 & 46.94 & 54.85 & 39.51 \\
            Parking:9 & [170,678] & 62.27 & 75.78 & 80.01 & 65.05 & \textbf{57.03} & 59.27 \\
            Parking:10 & [55,845] & \textbf{101.60} & 124.23 & 141.86 & 102.11 & 103.75 & 103.80 \\
            Parking:11 & [156,723] & 74.22 & 88.79 & 104.37 & 61.45 & \textbf{59.43} & 74.16 \\
            Parking:12 & [53,503] & 62.62 & 72.78 & 95.61 & 75.66 & 56.69 & \textbf{52.46} \\
            Parking:13 & [155,413] & \textbf{36.69} & 42.93 & 45.09 & 41.38 & 43.56 & 44.47 \\
            Parking:14 & [4,246] & 30.60 & 30.41 & 38.44 & 27.21 & \textbf{26.46} & 27.70 \\
            Parking:15 & [46,593] & \textbf{82.45} & 92.53 & 120.29 & 106.17 & 104.31 & 96.41 \\
            Parking:16 & [48,689] & 73.78 & 84.67 & 116.80 & 73.22 & \textbf{52.34} & 60.43 \\
            Parking:17 & [77,2811] & 307.67 & 361.59 & 451.15 & 299.45 & \textbf{236.23} & 268.42 \\
            Parking:18 & [1,847] & 63.88 & 79.55 & 84.51 & 77.57 & 60.53 & \textbf{56.34} \\
            Parking:19 & [1,696] & 57.90 & 73.52 & 79.15 & 71.45 & 54.61 & \textbf{47.50} \\
            Parking:20 & [452,1578] & 134.03 & 166.74 & 170.08 & 135.78 & 151.08 & \textbf{127.09} \\
            Parking:21 & [51,1534] & \textbf{113.36} & 153.19 & 145.35 & 138.09 & 142.10 & 123.52 \\
            Parking:22 & [524,3949] & 432.00 & 520.62 & 576.16 & 401.78 & 367.53 & \textbf{358.83} \\
            Parking:23 & [472,3429] & 317.78 & 462.98 & 533.55 & \textbf{313.17} & 349.44 & 362.69 \\
            Parking:24 & [331,1444] & \textbf{98.10} & 134.56 & 155.27 & 110.25 & 113.20 & 106.14 \\
            Parking:25 & [224,1023] & 87.47 & 105.46 & 109.22 & \textbf{80.14} & 109.43 & 100.09 \\
            Parking:26 & [390,1911] & \textbf{142.83} & 196.90 & 193.64 & 188.51 & 155.96 & 152.47 \\
            Parking:27 & [248,1561] & 155.64 & 211.38 & 228.86 & \textbf{145.20} & 158.50 & 170.15 \\
            AirQuality:0 & [-33,28] & \textbf{1.52} & 2.20 & 2.18 & 2.28 & 2.13 & 2.19 \\
            AirQuality:1 & [-19,41] & \textbf{1.32} & 1.78 & 1.83 & 1.77 & 1.78 & 2.00 \\
            AirQuality:2 & [991,1046] & \textbf{0.64} & 1.22 & 1.19 & 1.30 & 1.25 & 1.09 \\
            \bottomrule
        \end{tabular}
        \label{table:results_mae}
    \end{table}
\end{footnotesize}

\begin{footnotesize}
    \begin{table}[!t]
        \centering
        \caption{Sum of Borda counts of the models over the datasets. A higher value indicates more points in the voting score. The forward decoder RNN and backward decoder RNN in the proposed model are included as $\text{RNN}_\text{FW}$ and $\text{RNN}_\text{BW}$ respectively.}
        \setlength{\tabcolsep}{4pt}
        \begin{tabular}{lcccccc}
            \toprule
            & seq2seqImp & $\text{RNN}_\text{FW}$ & $\text{RNN}_\text{BW}$ & seq2seq  & RITS-I & BRITS-I \\
            \midrule
            MAE & \textbf{157} & 78 & 58 & 131 & 141 & 149 \\
            MRE & \textbf{157} & 74 & 56 & 138 & 144 & 145 \\
            \bottomrule
        \end{tabular}
        \label{table:results_borda}
    \end{table}
\end{footnotesize}

\section{Conclusion}

We propose a novel sequence-to-sequence model for recovering missing sensor data. Our decoder model merges two encoders that summarise the information of data before and after a missing data sequence. This is performed with a forward and backward RNN within the decoder. The decoder RNNs are merged together with a novel overarching output layer that performs scaling of the RNN cell outputs based on their proximity to observed data.

The proposed model is demonstrated on several time series datasets. It is shown that the proposed model produces the lowest errors in 12\% more cases than three other state-of-the-art models and is ranked as the best model according to the Borda count.

In future work, it is expected that significant improvement in the results could be achieved by using the attention mechanism \cite{bahdanau2014neural} between the encoders and the decoder. Furthermore, the scaling mechanism could possibly be improved by parameterising it within the model such that it can be learned. This could be achieved by using a softmax layer such as used in the attention mechanism.

\subsubsection*{Acknowledgments.}
The authors thank YiFan Zhang from CSIRO for the discussions around the topic of this study.
	
	\bibliographystyle{unsrt}
	\bibliography{Bibliography}

\end{document}